\documentclass[11pt]{article}
\usepackage{acl2016}
\usepackage{times}
\usepackage{url}
\usepackage{hyperref}
\usepackage{latexsym}

\usepackage{graphicx}
\usepackage[font=normalsize]{subfig}
\usepackage{algorithm}
\usepackage[noend]{algpseudocode}
\usepackage{multirow}

\usepackage{amsmath}
\usepackage{amssymb}
\usepackage{bm}
\usepackage{balance}
\usepackage{arydshln}
\usepackage{mathtools}

\aclfinalcopy 

\DeclareMathOperator*{\argmax}{arg\,max}

\title{Modeling Coverage for Neural Machine Translation}

\def\sstaff{$^\ddag$}
\def\fndaff{$^\dagger$}
\author{Zhaopeng Tu\fndaff ~~~ Zhengdong Lu\fndaff ~~~ Yang Liu\sstaff ~~~ Xiaohua Liu\fndaff ~~~ Hang Li\fndaff
\\
\\
{\fndaff {Noah's Ark Lab, Huawei Technologies, Hong Kong}}   \\
{\tt \{tu.zhaopeng,lu.zhengdong,liuxiaohua3,hangli.hl\}@huawei.com}\\
{\sstaff {Department of Computer Science and Technology, Tsinghua University, Beijing}}\\
{\tt liuyang2011@tsinghua.edu.cn}\\
}

\begin{document}

\maketitle

\begin{abstract}
\noindent Attention mechanism has enhanced state-of-the-art Neural Machine Translation (NMT) by jointly learning to align and translate. It tends to ignore past alignment information, however, which often leads to over-translation and under-translation. To address this problem, we propose coverage-based NMT in this paper. We maintain a coverage vector to keep track of the attention history. The coverage vector is fed to the attention model to help adjust future attention, which lets NMT system to consider more about untranslated source words. Experiments show that the proposed approach significantly improves both translation quality and alignment quality over standard attention-based NMT.\footnote{Our code is publicly available at \protect\url{https://github.com/tuzhaopeng/NMT-Coverage}.}
\end{abstract}

\section{Introduction}

The past several years have witnessed the rapid progress of end-to-end Neural Machine Translation (NMT) \cite{Sutskever:2014:NIPS,Bahdanau:2015:ICLR}. Unlike conventional Statistical Machine Translation (SMT) \cite{Koehn:2003:NAACL,Chiang:2007:CL}, NMT uses a single and large neural network to model the entire translation process. It enjoys the following advantages. First, the use of distributed representations of words can alleviate the curse of dimensionality \cite{Bengio:2003:JMLR}. Second, there is no need to explicitly design features to  capture translation regularities, which is quite difficult in SMT. Instead, NMT is capable of learning representations directly from the training data. Third, Long Short-Term Memory \cite{Hochreite:1997} enables NMT to capture long-distance reordering, which is a significant challenge in SMT.

\begin{figure*}[t]
\begin{center}
        \subfloat[Over-translation and under-translation generated by NMT.]{
            \includegraphics[width=0.4\textwidth]{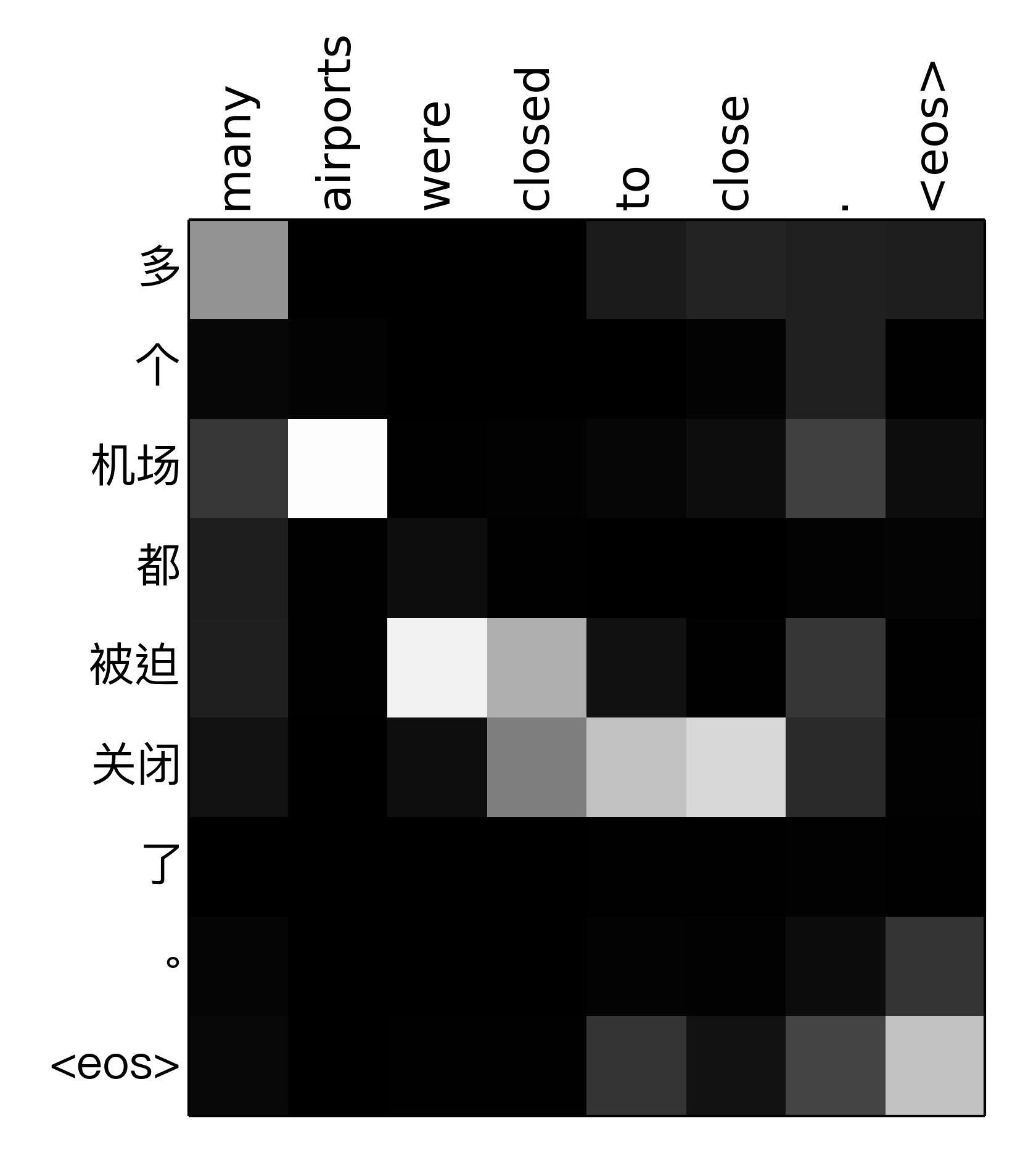}}\hspace{0.08\textwidth}
        \subfloat[Coverage model alleviates the problems of over-translation and under-translation.]{
            \includegraphics[width=0.45\textwidth]{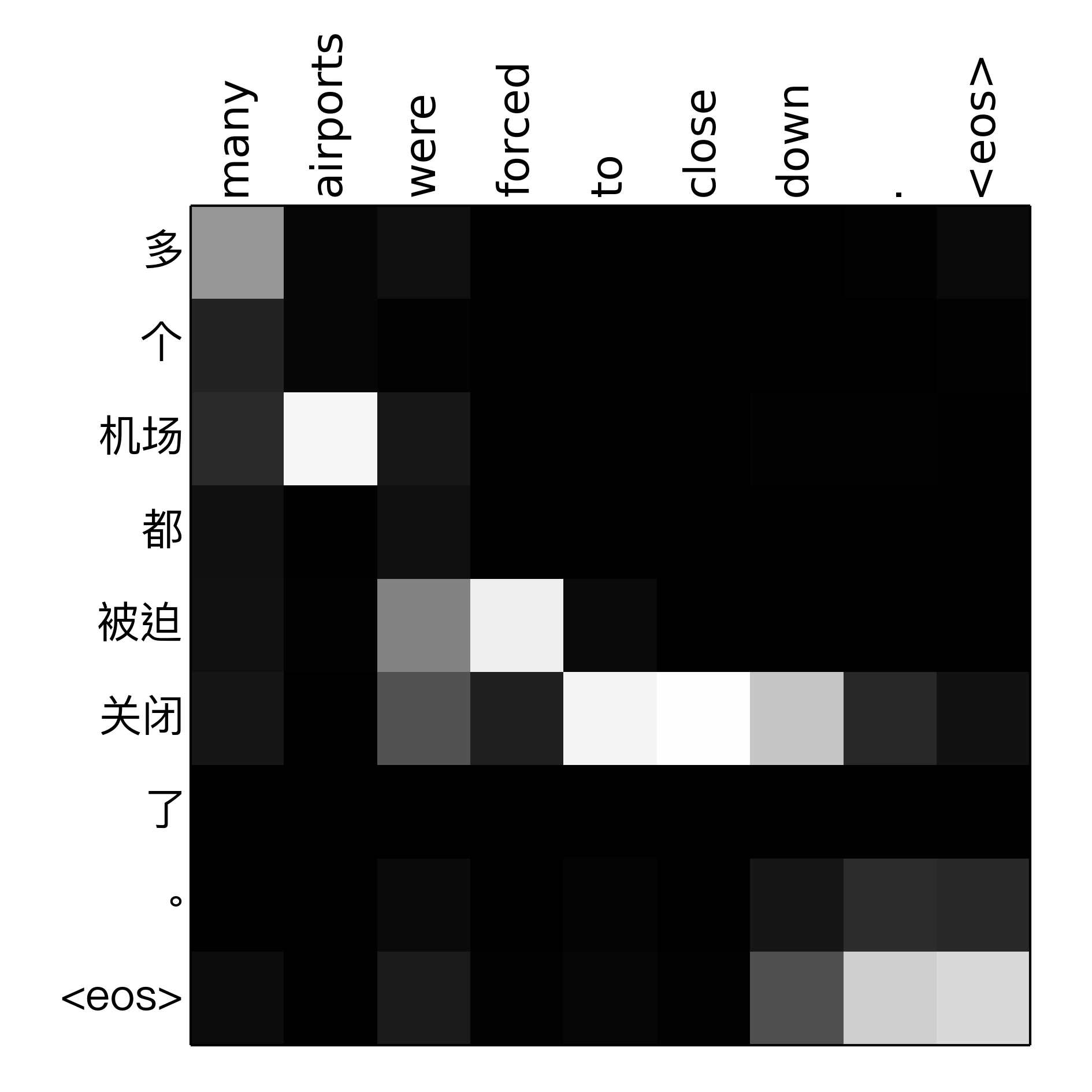}}
\end{center}
\caption{Example translations of (a) NMT without coverage, and (b) NMT with coverage. In conventional NMT without coverage, the Chinese word ``\emph{gu{\= a}nb{\`\i}}'' is over translated to ``\emph{close}(\emph{d})'' twice, while ``\emph{b{\` e}ip{\` o}}'' (means ``\emph{be forced to}'') is mistakenly untranslated. Coverage model alleviates these problems by tracking the ``coverage'' of source words.}
\label{figure-examples}
\end{figure*}

NMT has a serious problem, however, namely lack of \emph{coverage}. In phrase-based SMT \cite{Koehn:2003:NAACL}, a decoder maintains a coverage vector to indicate whether a source word is translated or not.  This is important for ensuring that each source word is translated in decoding. The decoding process is completed when all source words are ``covered'' or translated. In NMT, there is no such coverage vector and the decoding process ends only when the end-of-sentence mark is produced. We believe that lacking coverage might result in the following problems in conventional NMT:
\begin{enumerate}
\item Over-translation: some words are unnecessarily translated for multiple times;
\item Under-translation: some words are mistakenly untranslated.
\end{enumerate}
Specifically, in the state-of-the-art attention-based NMT model \cite{Bahdanau:2015:ICLR}, generating a target word heavily depends on the relevant parts of the source sentence, and a source word is involved in generation of all target words. As a result, over-translation and under-translation inevitably happen because of ignoring the ``coverage'' of source words (i.e., number of times a source word is translated to a target word). Figure~\ref{figure-examples}(a) shows an example: the Chinese word ``\emph{gu{\= a}nb{\`\i}}'' is over translated to ``\emph{close}(\emph{d})'' twice, while ``\emph{b{\` e}ip{\` o}}'' (means ``\emph{be forced to}'') is mistakenly untranslated.

In this work, we propose a coverage mechanism to NMT (\textsc{NMT-Coverage}) to alleviate the over-translation and under-translation problems. Basically, we append a coverage vector to the intermediate representations of an NMT model, which are sequentially updated after each attentive read during the decoding process, to keep track of the attention history. The coverage vector, when entering into attention model, can help adjust the future attention and significantly improve the overall alignment between the source and target sentences. This design contains many particular cases for coverage modeling with contrasting characteristics,  which all share a clear linguistic intuition and yet can be trained in a data driven fashion. 
Notably, we achieve significant improvement even by simply using the sum of previous alignment probabilities as coverage for each word, as a successful example of incorporating linguistic knowledge into neural network based NLP models. 

Experiments show that \textsc{NMT-Coverage} significantly outperforms conventional attention-based NMT on both translation and alignment tasks.
Figure~\ref{figure-examples}(b) shows an example, in which \textsc{NMT-Coverage} alleviates the over-translation and under-translation problems that NMT without coverage suffers from.

\section{Background}

\begin{figure}[t]
\centering
\includegraphics[width=0.45\textwidth]{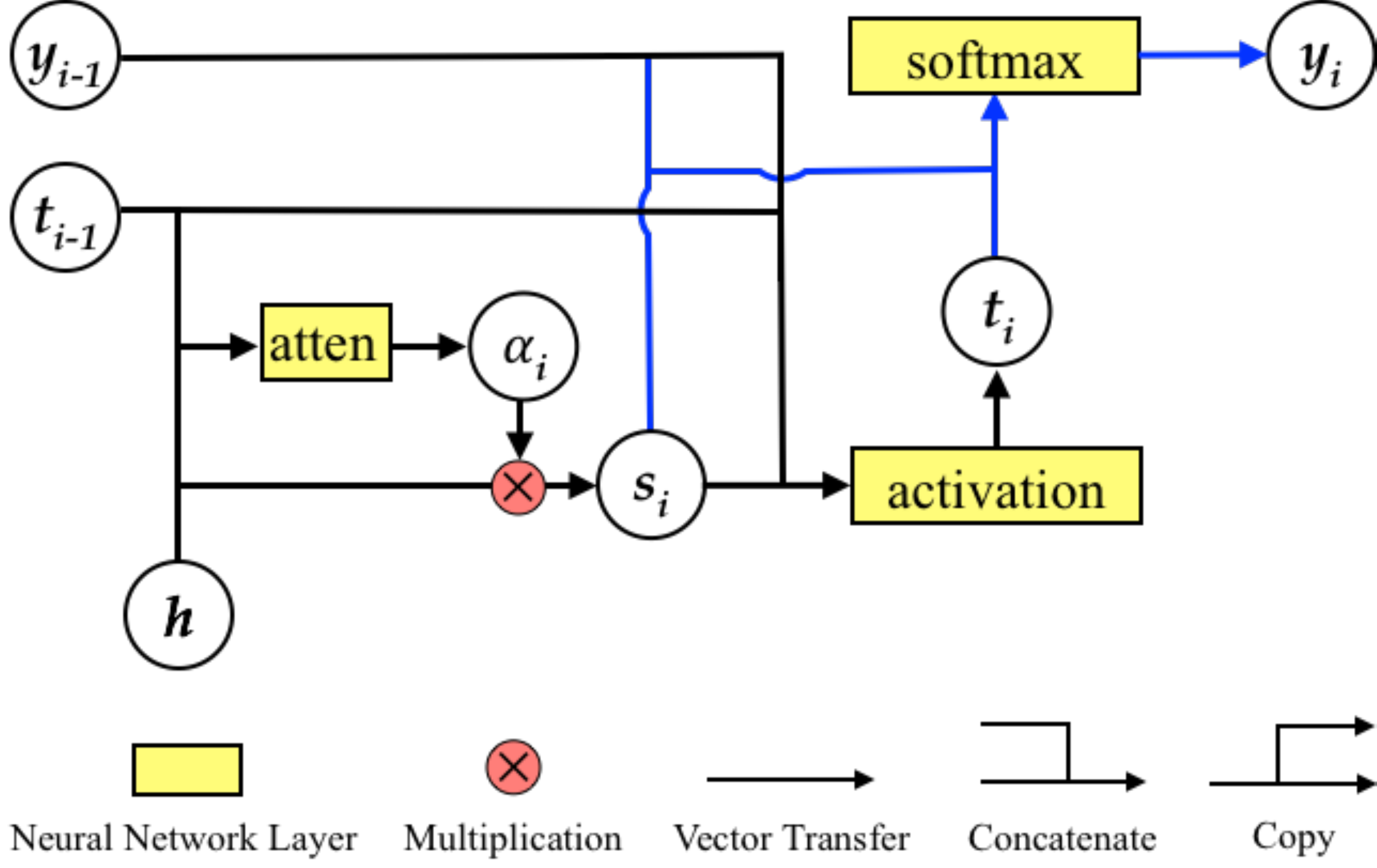}
\caption{Architecture of attention-based NMT. Whenever possible, we omit the source index $j$ to make the illustration less cluttered.}
\label{figure-nmt}
\end{figure}

Our work is built on attention-based NMT~\cite{Bahdanau:2015:ICLR}, which simultaneously conducts dynamic alignment and generation of the target sentence, as illustrated in Figure~\ref{figure-nmt}. It produces the translation by generating one target word $y_i$ at each time step.
Given an input sentence ${\bf x}=\{x_1, \dots, x_{J}\}$ and previously generated words $\{y_1, \dots, y_{i-1}\}$, the probability of generating next word $y_i$ is
\begin{equation}
P(y_i|y_{<i}, {\bf x}) = softmax\big(g(y_{i-1}, {\bf t}_i, {\bf s}_i)\big)
\label{eqn-nmt-prediction}
\end{equation}
where $g$ is a non-linear function, and ${\bf t}_i$ is a decoding state for time step $i$, computed by
\begin{equation}
{\bf t}_i = f({\bf t}_{i-1}, y_{i-1}, {\bf s}_i)
\label{eqn-nmt-state}
\end{equation}
Here the activation function $f(\cdot)$ is a Gated Recurrent Unit (GRU)~\cite{Cho:2014:EMNLP}, and ${\bf s}_i$ is a distinct source representation for time $i$, calculated as a weighted sum of the source annotations:
\begin{equation}
{\bf s}_i = \sum_{j=1}^{J}{\alpha_{i,j}\cdot {\bf h}_j}
\label{eqn-context}
\end{equation}
where ${\bf h}_j={[\overrightarrow{h}_j^{\top};\overleftarrow{h}_j^{\top}]}^\top$ is the annotation of $x_j$ from a bi-directional Recurrent Neural Network (RNN)~\cite{Schuster:1997:TSP}, and its weight $\alpha_{i,j}$ is computed by
\begin{equation}
\alpha_{i,j} = \frac{\exp(e_{i,j})}{\sum_{k=1}^{J} \exp(e_{i,k})} 
\label{eqn-alignment-probability} 
\end{equation}
and 
\begin{flalign}
e_{i,j} &= a({\bf t}_{i-1}, {\bf h}_j) \nonumber \\
          &= v_a^{\top} \tanh (W_a {\bf t}_{i-1} + U_a {\bf h}_j)
\label{eqn-alignment-model}
\end{flalign}
is an \emph{attention model} that scores how well $y_i$ and ${\bf h}_j$ match.
With the attention model, it avoids the need to represent the entire source sentence with a single vector. Instead, the decoder selects parts of the source sentence to pay attention to, thus exploits an \emph{expected annotation} ${\bf s}_i$ over possible alignments $\alpha_{i,j}$ for each time step $i$. 

However, the attention model fails to take advantage of past alignment information, which is found useful to avoid over-translation and under-translation problems in conventional SMT~\cite{Koehn:2003:NAACL}. For example, if a source word is translated in the past, it is less likely to be translated again and should be assigned a lower alignment probability.

\section{Coverage Model for NMT}
\label{sec-coverage-model}

\begin{figure}[t]
\centering
\includegraphics[width=0.4\textwidth]{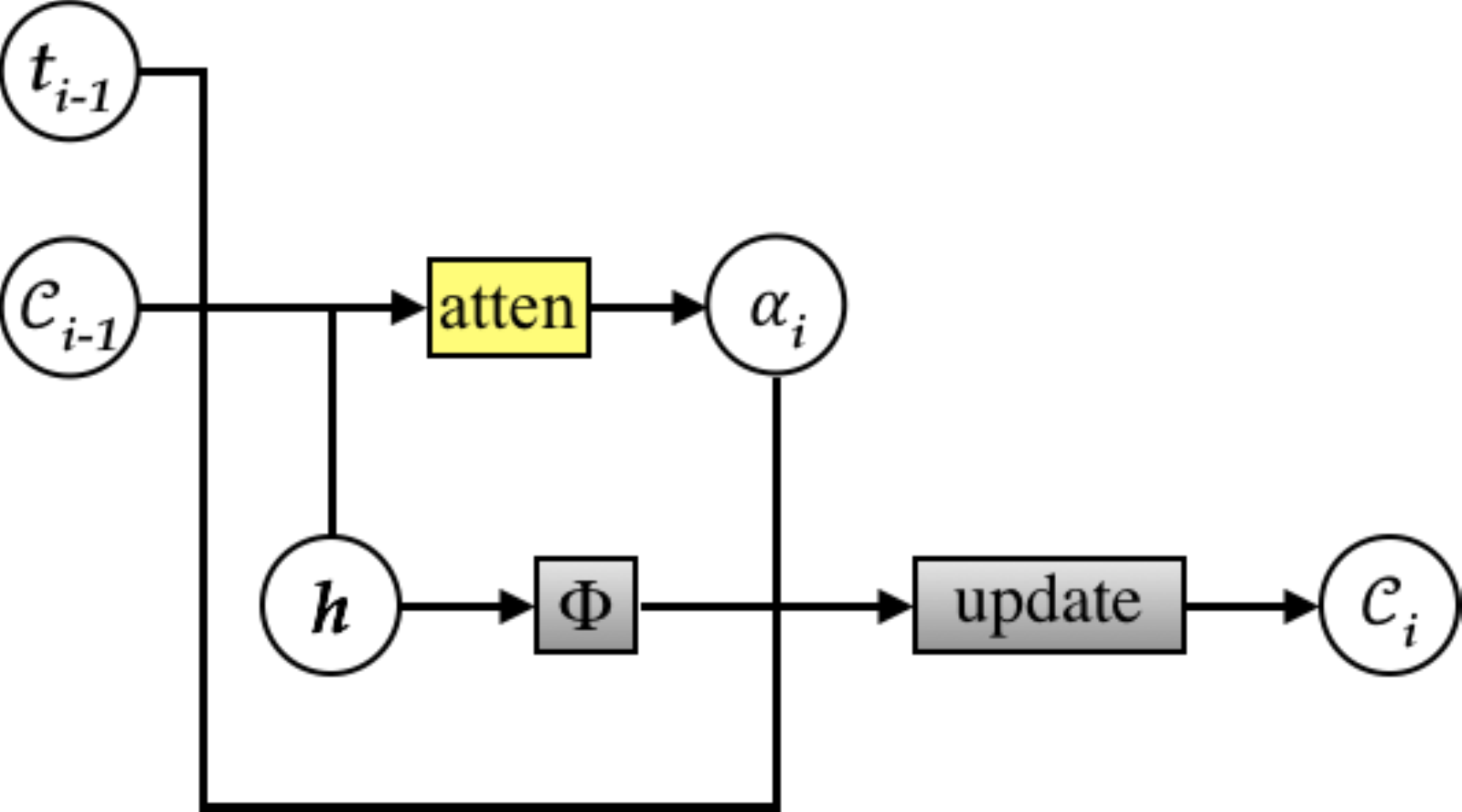}
\caption{Architecture of coverage-based attention model. A coverage vector ${\mathcal{C}}_{i-1}$ is maintained to keep track of which source words have been translated before time $i$. 
Alignment decisions $\alpha_{i}$ are made jointly taking into account past alignment information embedded in $\mathcal{C}_{i-1}$, which lets the attention model to consider more about untranslated source words.}
\label{figure-coverage-alignment}
\end{figure}

In SMT, a coverage set is maintained to keep track of which source words have been translated (``covered'') in the past. Let us take ${\bf x}=\{x_1, x_2, x_3, x_4\}$ as an example of input sentence. The initial coverage set is $\mathcal{C}=\{0, 0, 0, 0\}$ which denotes that no source word is yet translated. When a translation rule $bp = (x_2 x_3, y_my_{m+1})$ is applied, we produce one hypothesis labelled with coverage $\mathcal{C}=\{0, 1, 1, 0\}$. It means that the second and third source words are translated. The goal is to generate translation with full coverage $\mathcal{C}=\{1, 1, 1, 1\}$. 
A source word is translated when it is covered by one translation rule, and it is not allowed to be translated again in the future  (i.e., \emph{hard coverage}).
In this way, each source word is guaranteed to be translated and only be translated once.
As shown, coverage is essential for SMT since it avoids gaps and overlaps in translation of source words.

Modeling coverage is also important for attention-based NMT models, since they generally lack a mechanism to indicate whether a certain source word has been translated, and therefore are prone to the  ``coverage'' mistakes: some parts of source sentence have been translated more than once or not translated. For NMT models, directly modeling coverage is less straightforward, but the problem can be significantly alleviated by keeping track of the attention signal during the decoding process. The most natural way for doing that would be to append a coverage vector to the annotation of each source word (i.e., ${\bf h}_j$), which is initialized as a zero vector but updated after every attentive read of the corresponding annotation. 
The coverage vector is fed to the attention model to help adjust future attention, which lets NMT system to consider more about untranslated source words, as illustrated in Figure~\ref{figure-coverage-alignment}.

\subsection{Coverage Model}

Since the coverage vector summarizes the attention record for ${\bf h}_j$ (and therefore for a small neighbor centering at the $j^{th}$ source word), it will discourage further attention to it if it has been heavily attended, and implicitly push the attention to the less attended segments of the source sentence since the attention weights are normalized to one. This can potentially solve both coverage mistakes mentioned above, when modeled and learned properly.

Formally, the coverage model is given by 
\begin{equation}
\mathcal{C}_{i,j} = g_{update}\big(\mathcal{C}_{i-1,j}, \alpha_{i,j}, \Phi({\bf h}_j), \Psi\big)
\label{eqn-coverage-general}
\end{equation}
where
\begin{itemize}
  \item $g_{update}(\cdot)$ is the function that updates $\mathcal{C}_{i,j}$ after the new attention $\alpha_{i,j}$ at time step $i$ in the decoding process;
  \item $\mathcal{C}_{i,j}$ is a $d$-dimensional coverage vector summarizing the history of attention till time step $i$ on ${\bf h}_j$;
  \item $\Phi({\bf h}_j)$ is a word-specific feature with its own parameters;
  \item $\Psi$ are auxiliary inputs exploited in different sorts of coverage models.
\end{itemize}
Equation~\ref{eqn-coverage-general} gives a rather general model, which could take different function forms for $g_{update}(\cdot)$ and $\Phi(\cdot)$, and different auxiliary inputs $\Psi$ (e.g., previous decoding state ${\bf t}_{i-1}$). In the rest of this section, we will give a number of representative implementations of the coverage model, which either leverage more linguistic information (Section~\ref{sec-linear-coverage}) or resort to the flexibility of neural network approximation (Section~\ref{sec-nonlinear-coverage}).

\subsubsection{Linguistic Coverage Model}\label{sec-linear-coverage}

We first consider at linguistically inspired model which has a small number of parameters, as well as clear interpretation.
While the linguistically-inspired coverage in NMT is similar to that in SMT, there is one key difference: 
it indicates what percentage of source words have been translated (i.e., \emph{soft coverage}). 
In NMT, each target word $y_i$ is generated from all source words with probability $\alpha_{i,j}$ for source word $x_j$. 
In other words, the source word $x_j$ is involved in generating all target words and the probability of generating target word $y_i$ at time step $i$ is $\alpha_{i,j}$.
Note that unlike in SMT in which each source word is \emph{fully translated} at one decoding step, the source word $x_j$ is \emph{partially translated} at each decoding step in NMT.
Therefore, the coverage at time step $i$ denotes the translated ratio of that each source word is translated.

We use a scalar ($d=1$) to represent linguistic coverage for each source word and employ an accumulate operation for $g_{update}$. The initial value of linguistic coverage is zero, which denotes that the corresponding source word is not translated yet. We iteratively construct linguistic coverages through accumulation of alignment probabilities generated by the attention model, each of which is normalized by a distinct context-dependent weight.
The coverage of source word $x_j$ at time step $i$ is computed by
\begin{equation}
\mathcal{C}_{i,j}= \mathcal{C}_{i-1,j} + \frac{1}{\Phi_j} \alpha_{i,j} = \frac{1}{\Phi_j} \sum_{k=1}^{i} \alpha_{k,j}
\label{eqn-fertility-coverage}
\end{equation}
where $\Phi_j$ is a pre-defined weight which indicates the number of target words $x_j$ is expected to generate. 
The simplest way is to follow Xu et al.~\shortcite{Xu:2015:ICML} in image-to-caption translation to fix $\Phi=1$ for all source words, which means that we directly use the sum of previous alignment probabilities without normalization as coverage for each word, as done in~\cite{Cohn:2016:NAACL}.

However, in machine translation, different types of source words may contribute differently to the generation of target sentence. Let us take the sentence pairs in Figure~\ref{figure-examples} as an example. The noun in the source sentence ``\emph{j{\=\i}ch{\v a}ng}'' is translated into one target word ``\emph{airports}'', while the adjective ``\emph{b{\` e}ip{\` o}}'' is translated into three words ``\emph{were forced to}''. Therefore, we need to assign a distinct $\Phi_j$ for each source word.
Ideally, we expect $\Phi_j = \sum_{i=1}^{I} \alpha_{i,j}$ with $I$ being the total number of time steps in decoding. However, such desired value is not available before decoding, thus is not suitable in this scenario. 

\paragraph{Fertility}
To predict $\Phi_j$, 
we introduce the concept of \emph{fertility}, which is firstly proposed in word-level SMT~\cite{Brown:1993:CL}. 
Fertility of source word $x_j$ tells how many target words $x_j$ produces. In SMT, the fertility is a random variable $\Phi_j$, whose distribution $p(\Phi_j=\phi)$ is determined by the parameters of word alignment models (e.g., IBM models).
In this work, we simplify and adapt fertility from the original model and compute the fertility $\Phi_j$ by\footnote{Fertility in SMT is a random variable with a set of fertility probabilities, $n(\Phi_j|x_j) = p(\Phi_{<j}, {\bf x})$, which depends on the fertilities of previous source words. To simplify the calculation and adapt it to the attention model in NMT, we define the fertility in NMT as a constant number,  which is independent of previous fertilities.}
\begin{equation}
\Phi_{j} = \mathcal{N} (x_j | {\bf x}) = N \cdot \sigma(U_f {\bf h}_j)
\label{eqn-fertility}
\end{equation}
where $N \in \mathbb{R}$ is a predefined constant to denote the maximum number of target words one source word can produce, $\sigma(\cdot)$ is a logistic sigmoid function, and $U_f \in \mathbb{R}^{1 \times 2n}$ is the weight matrix.
Here we use ${\bf h}_j$ to denote $(x_j|{\bf x})$ since ${\bf h}_j$ contains information about the whole input sentence with a strong focus on the parts surrounding $x_j$~\cite{Bahdanau:2015:ICLR}.
Since $\Phi_j$ does not depend on $i$, we can pre-compute it before decoding to minimize the computational cost.

\subsubsection{Neural Network Based Coverage Model} \label{sec-nonlinear-coverage}

\begin{figure}[t]
\centering
\includegraphics[width=0.25\textwidth]{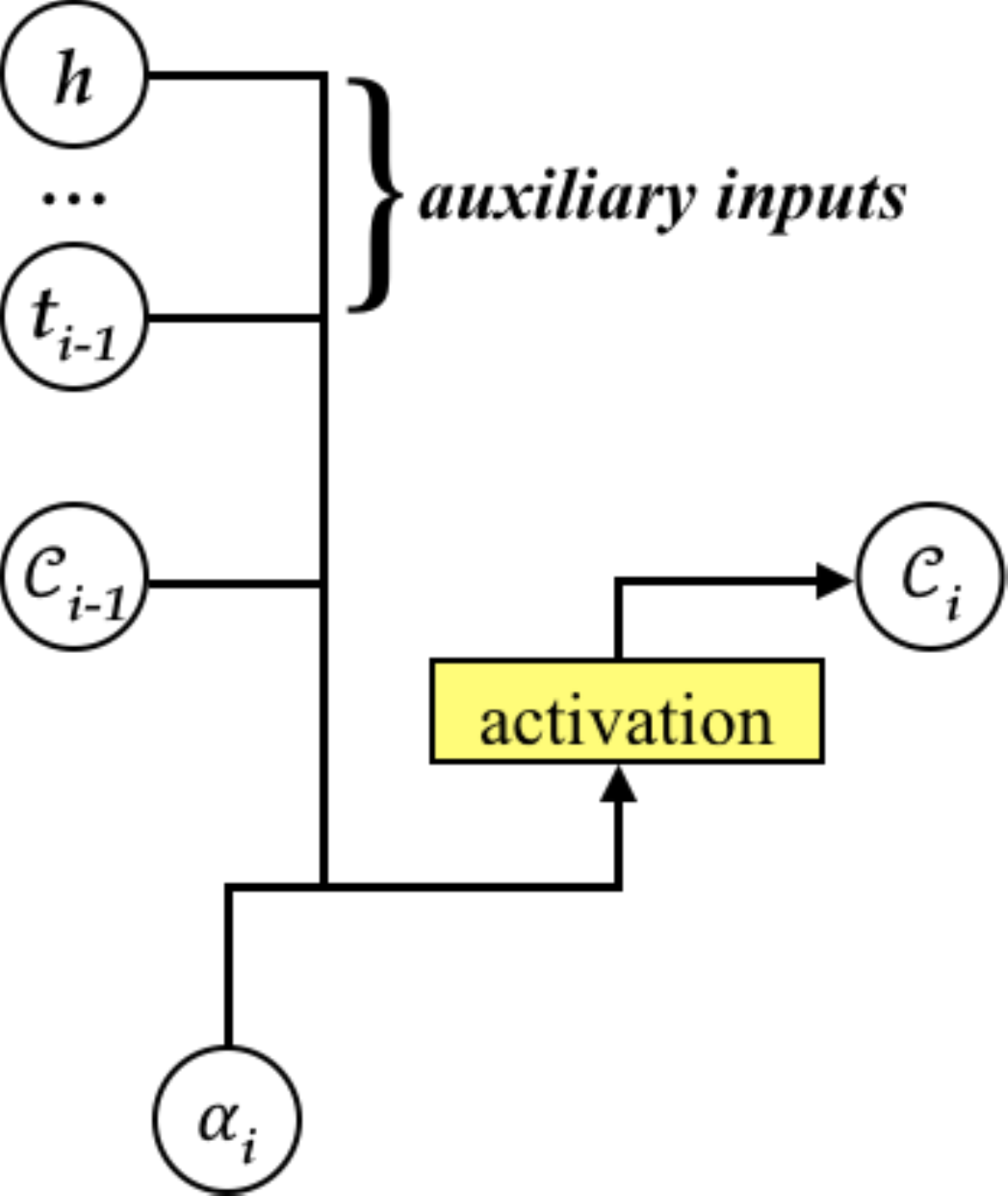}
\caption{NN-based coverage model.}
\label{figure-recurrent}
\end{figure}

We next consider Neural Network (NN) based coverage model.
When $\mathcal{C}_{i,j}$ is a vector ($d>1$) and $g_{update}(\cdot)$  is a neural network, we actually have an RNN model for coverage, as illustrated in Figure~\ref{figure-recurrent}. In this work, we take the following form:
\begin{flalign}
\mathcal{C}_{i,j} &= f (\mathcal{C}_{i-1,j}, \alpha_{i,j}, {\bf h}_j, {\bf t}_{i-1}) \nonumber
\end{flalign}
where $f(\cdot)$ is a nonlinear activation function and ${\bf t}_{i-1}$ is the auxiliary input that encodes past translation information.
Note that we leave out the word-specific feature function $\Phi(\cdot)$ and only take the input annotation ${\bf h}_j$ as the input to the coverage RNN. It is important to emphasize that the NN-based coverage model is able to be fed with arbitrary inputs, such as the previous attentional context ${\bf s}_{i-1}$. Here we only employ $\mathcal{C}_{i-1,j}$ for past alignment information, ${\bf t}_{i-1}$ for past translation information, and ${\bf h}_j$ for word-specific bias.\footnote{In our preliminary experiments, considering more inputs (e.g., current and previous attentional contexts, unnormalized attention weights $e_{i,j}$) does not always lead to better translation quality. Possible reasons include: 1) the inputs contains duplicate information, and 2) more inputs introduce more back-propagation paths and therefore make it difficult to train. In our experience, one principle is to only feed the coverage model inputs that contain distinct information, which are complementary to each other.}

\paragraph{Gating}
The neural function $f(\cdot)$ can be either a simple activation function $\tanh$ or a gating function that proves useful to capture long-distance dependencies. 
In this work, we adopt GRU for the gating activation since it is simple yet powerful~\cite{Chung:2014:arXiv}. Please refer to~\cite{Cho:2014:EMNLP} for more details about GRU.

\vspace{5pt}
\paragraph{Discussion}
Intuitively, the two types of models summarize coverage information in ``different languages''. Linguistic models summarize coverage information in human language, which has a clear interpretation to humans. Neural models encode coverage information in ``neural language'', which can be ``understood'' by neural networks and let them to decide how to make use of the encoded coverage information.

\subsection{Integrating Coverage into NMT}
\label{sec-coverage-alignment-model}

Although attention based model has the capability of jointly making alignment and translation, it does not take into consideration translation history. Specifically, a source word that has significantly contributed to the generation of target words in the past, should be assigned lower alignment probabilities, which may not be the case in attention based NMT. To address this problem, we propose to calculate the alignment probabilities by incorporating past alignment information embedded in the coverage model.

Intuitively, at each time step $i$ in the decoding phase, coverage from time step ($i-1$) serves as an additional input to the attention model, which provides complementary information of that how likely the source words are translated in the past.
We expect the coverage information would guide the attention model to focus more on untranslated source words (i.e., assign higher alignment probabilities). In practice, we find that the coverage model does fulfill the expectation (see Section~\ref{sec-experiments}). The translated ratios of source words from linguistic coverages negatively correlate to the corresponding alignment probabilities. 

More formally, we rewrite the attention model in Equation~\ref{eqn-alignment-model} as
\begin{flalign}
 e_{i,j} &= a({\bf t}_{i-1}, {\bf h}_j, \mathcal{C}_{i-1, j})   \nonumber \\
& = v_a^{\top} \tanh (W_a {\bf t}_{i-1} + U_a {\bf h}_j + V_a \mathcal{C}_{i-1, j}) \nonumber
\end{flalign}
where $\mathcal{C}_{i-1,j}$ is the coverage of source word $x_j$ before time $i$. $V_a \in \mathbb{R}^{n\times d}$ is the  weight matrix for coverage with $n$ and $d$ being the numbers of hidden units and coverage units, respectively.

\section{Training}
\label{sec-training}
We take end-to-end learning for the \textsc{NMT-Coverage} model, which learns not only the parameters for the ``original" NMT (i.e., $\theta$ for encoding RNN, decoding RNN, and attention model) but also the parameters for coverage modeling (i.e., $\eta$ for annotation and guidance of attention) . More specifically, we choose to maximize the likelihood of reference sentences as most other NMT models (see, however~\cite{Shen:2016:ACL}):
\begin{flalign}
(\theta^*, \eta^{*}) = \argmax_{\theta, \eta}\sum_{n=1}^{N} \log P({\bf y}_n|{\bf x}_n; \theta, \eta)
\label{eqn-training}
\end{flalign}

\paragraph{No auxiliary objective}
For the coverage model with a clearer linguistic interpretation (Section \ref{sec-linear-coverage}), it is possible to inject an auxiliary objective function on some intermediate representation. More specifically, we may have the following objective:
\begin{flalign}
(\theta^*, \eta^{*}) = & \argmax_{\theta, \eta} \sum_{n=1}^{N} \Bigg\{\log P({\bf y}_n|{\bf x}_n; \theta, \eta)  \nonumber \\
                                       & - \lambda \Big\{\sum_{j=1}^{J}(\Phi_j - \sum_{i=1}^{I}{\alpha_{i,j}})^2; \eta\Big\} \Bigg\} \nonumber
\label{eqn-coverage-training}
\end{flalign}
where the term $\big\{\sum_{j=1}^{J}(\Phi_j - \sum_{i=1}^{I}{\alpha_{i,j}})^2; \eta\big\}$  penalizes the discrepancy between the sum of alignment probabilities and the expected fertility for linguistic coverage. This is similar to the more explicit training for fertility as in  Xu et al.~\shortcite{Xu:2015:ICML}, which 
encourages the model to pay equal attention to every part of the image (i.e., $\Phi_j=1$).
However, our empirical study shows that the combined objective consistently worsens the translation quality while slightly improves the alignment quality. 

Our training strategy poses less constraints on the dependency between $\Phi_j$ and the attention than a more explicit strategy taken in~\cite{Xu:2015:ICML}. We let the objective associated with the translation quality (i.e., the likelihood) to drive the training, as in Equation~\ref{eqn-training}. This strategy is arguably advantageous, since the attention weight on a hidden state ${\bf h}_j$ cannot be interpreted as the proportion of the corresponding word being translated in the target sentence. For one thing, the hidden state ${\bf h}_j$, after the transformation from encoding RNN, bears the contextual information from other parts of the source sentence, and thus loses the rigid correspondence with the corresponding word. Therefore, penalizing the discrepancy between the sum of alignment probabilities and the expected fertility does not hold in this scenario.

\section{Experiments}
\label{sec-experiments}

\begin{table*}[t]
\centering
\begin{tabular}{c|l|l|llll}
    \#	& {\bf System}	&	{\bf \#Params}		&	{\bf MT05}  &  {\bf MT06}	  &	{\bf MT08}  &  {\bf Avg.}\\
    \hline
    1	&	Moses	              					&	--	&	31.37	&	30.85	&	23.01	&	28.41\\
    \hline
    2	&	GroundHog						&	84.3M	&	30.61	&	31.12	&	23.23	&	28.32\\
    \hdashline
    3	&	+ Linguistic coverage w/o fertility             	&	+1K	&	31.26$^\dagger$	&	32.16$^\dagger$$^\ddag$		&	24.84$^\dagger$$^\ddag$		&	29.42\\
    4	&	+ Linguistic coverage w/ fertility		&	+3K	&	32.36$^\dagger$$^\ddag$		&	32.31$^\dagger$$^\ddag$		&	24.91$^\dagger$$^\ddag$		&	29.86\\
    \hdashline
    5	&	+ NN-based coverage w/o gating ($d=1$)		&	+4K	&	31.94$^\dagger$$^\ddag$	&	32.11$^\dagger$$^\ddag$		&	23.31	&	29.12\\
    6	&	+ NN-based coverage w/ gating ($d=1$)           	&	+10K	&	31.94$^\dagger$$^\ddag$		&	32.16$^\dagger$$^\ddag$		&	24.67$^\dagger$$^\ddag$		&	29.59\\
    7	&	+ NN-based coverage w/ gating ($d=10$)           	&	+100K	&	{\bf 32.73}$^\dagger$$^\ddag$	&	{\bf 32.47}$^\dagger$$^\ddag$		&	{\bf 25.23}$^\dagger$$^\ddag$		&	{\bf 30.14}\\
\end{tabular}
\caption{Evaluation of translation quality. $d$ denotes the dimension of NN-based coverages, and $\dagger$ and $\ddag$  indicate statistically significant difference ($p < 0.01$) from GroundHog and Moses, respectively. ``+'' is on top of the baseline system GroundHog.}
\label{table-translation-results}
\end{table*}

\subsection{Setup}

We carry out experiments on a Chinese-English translation task. 
Our training data for the translation task consists of 1.25M sentence pairs extracted from LDC corpora\footnote{The corpora include LDC2002E18, LDC2003E07, LDC2003E14, Hansards portion of LDC2004T07, LDC2004T08 and LDC2005T06.}
, with 27.9M Chinese words and 34.5M English words respectively.
We choose NIST 2002 dataset as our development set, and the NIST 2005, 2006 and 2008 datasets as our test sets.
We carry out experiments of the alignment task on the evaluation dataset from~\cite{Liu:2015:AAAI}, which contains 900 manually aligned Chinese-English sentence pairs. 
We use the case-insensitive 4-gram NIST BLEU score~\cite{Papineni:2002} for the translation task, and the alignment error rate (AER)~\cite{Och:2003} for the alignment task.
To better estimate the quality of the soft alignment probabilities generated by NMT, we propose a variant of AER, naming \emph{SAER}:
\begin{eqnarray}
SAER=1-\frac{|M_{A} \times M_{S}|+|M_{A} \times M_{P}|}{|M_{A}|+|M_{S}|} \nonumber
\end{eqnarray}
where $A$ is a candidate alignment, and  $S$ and $P$ are the sets of sure and possible links in the reference alignment respectively ($S\subseteq P$).
$M$ denotes alignment matrix, and for both $M_{S}$ and $M_{P}$ we assign the elements that correspond to the existing links in $S$ and $P$ with probabilities $1$ while assign the other elements with probabilities $0$. 
In this way, we are able to better evaluate the quality of the soft alignments produced by attention-based NMT.
We use \emph{sign-test}~\cite{Collins:2005} for statistical significance test.

For efficient training of the neural networks, we limit the source and target vocabularies to the most frequent 30K words in Chinese and English, covering approximately 97.7\% and 99.3\% of the two corpora respectively. 
All the out-of-vocabulary words are mapped to a special token \texttt{\small UNK}.
We set $N=2$ for the fertility model in the linguistic coverages. 
We train each model with the sentences of length up to 80 words in the training data. The word embedding dimension is 620 and the size of a hidden layer is 1000. 
All the other settings are the same as in~\cite{Bahdanau:2015:ICLR}.

We compare our method with two state-of-the-art models of SMT and NMT\footnote{There are recent progress on aggregating multiple models or enlarging the vocabulary(e.g., in ~\cite{Jean:2015:ACL}), but here we focus on the generic models.}:
\begin{itemize}
    \item {\bf Moses}~\cite{Koehn:2007:ACL}: an open source phrase-based translation system with default configuration and a 4-gram language model trained on the target portion of training data.
    \item {\bf GroundHog}~\cite{Bahdanau:2015:ICLR}:  an attention-based NMT system.
\end{itemize}

\subsection{Translation Quality}

Table~\ref{table-translation-results} shows the translation performances measured in BLEU score. 
Clearly the proposed \textsc{NMT-Coverage} significantly improves the translation quality in all cases, although there are still considerable differences among different variants.
\paragraph{Parameters} Coverage model introduces few parameters. The baseline model (i.e., GroundHog) has 84.3M parameters. The linguistic coverage using fertility introduces 3K parameters (2K for fertility model), and the NN-based coverage with gating introduces 10K$\times d$ parameters (6K$\times d$ for gating), where $d$ is the dimension of the coverage vector. In this work, the most complex coverage model only introduces 0.1M additional parameters, which is quite small compared to the number of parameters in the existing model (i.e., 84.3M).

\paragraph{Speed} Introducing the coverage model slows down the training speed, but not significantly. When running on a single GPU device Tesla K80, the speed of the baseline model is 960 target words per second. System 4 (``+Linguistic coverage with fertility'') has a speed of 870 words per second, while System 7 (``+NN-based coverage (d=10)'') achieves a speed of 800 words per second.

\paragraph{Linguistic Coverages} (Rows 3 and 4): Two observations can be made. First,  the simplest linguistic coverage (Row 3) already significantly improves translation performance by 1.1 BLEU points, indicating that coverage information is very important to the attention model. Second, incorporating fertility model boosts the performance by better estimating the covered ratios of source words.

\paragraph{NN-based Coverages} (Rows 5-7): (1) \emph{Gating} (Rows 5 and 6): Both variants of NN-based coverages outperform GroundHog with averaged gains of 0.8 and 1.3 BLEU points, respectively. Introducing gating activation function improves the performance of coverage models, which is consistent with the results in other tasks~\cite{Chung:2014:arXiv}. (2) \emph{Coverage dimensions} (Rows 6 and 7): Increasing the dimension of coverage models further improves the translation performance by 0.6 point in BLEU score, at the cost of introducing more parameters (e.g., from 10K to 100K).\footnote{In a pilot study, further increasing the coverage dimension only slightly improved the translation performance. One possible reason is that encoding the relatively simple coverage information does not require too many dimensions.}

\begin{table*}[t]
\centering
\begin{tabular}{l|cc|cc}
    {\bf Model}		&	{\bf Adequacy}	&	{\bf Fluency}	&	\parbox[c][24pt]{50pt}{\bf Under-Translation}	&	\parbox[c]{50pt}{\bf Over-Translation}\\
    \hline
    GroundHog			&	3.06	&	3.54	&	25.0\%	&	4.5\%\\
    + NN cov. w/ gating {\small ($d=10$)}				&	3.28	&	3.73	&	16.7\%	&	2.7\%\\
\end{tabular}
\caption{Subjective evaluation of translation adequacy and fluency. The numbers in the last two columns denote the percentages of source words are under-translated and over-translated, respectively.}
\label{table-subjective-evaluation}
\end{table*}

\begin{figure*}[t]
\begin{center}
        \subfloat[Groundhog]{
            \includegraphics[width=0.45\textwidth]{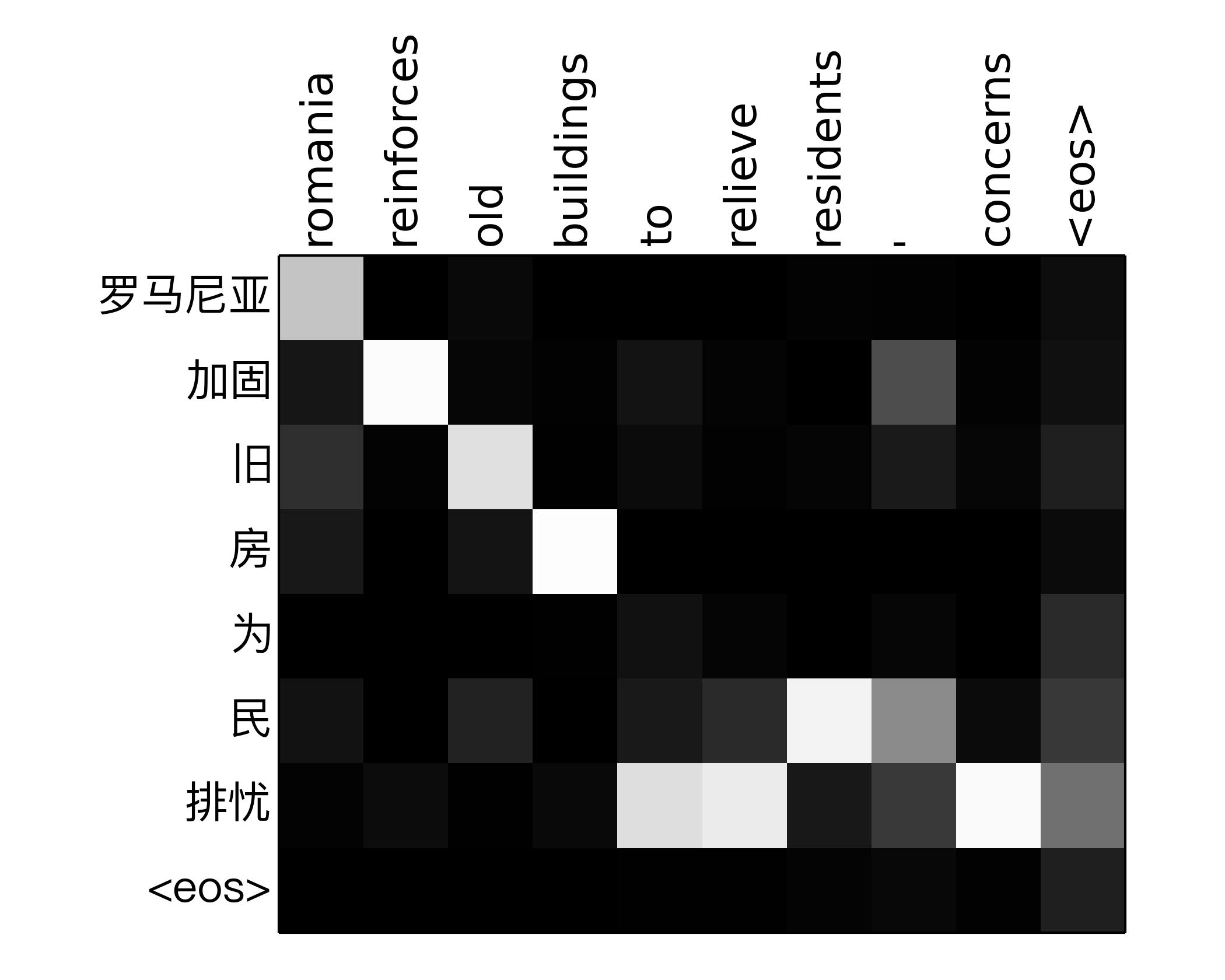}} \hspace{0.08\textwidth}
        \subfloat[+ NN cov. w/ gating {\small ($d=10$)}]{
            \includegraphics[width=0.45\textwidth]{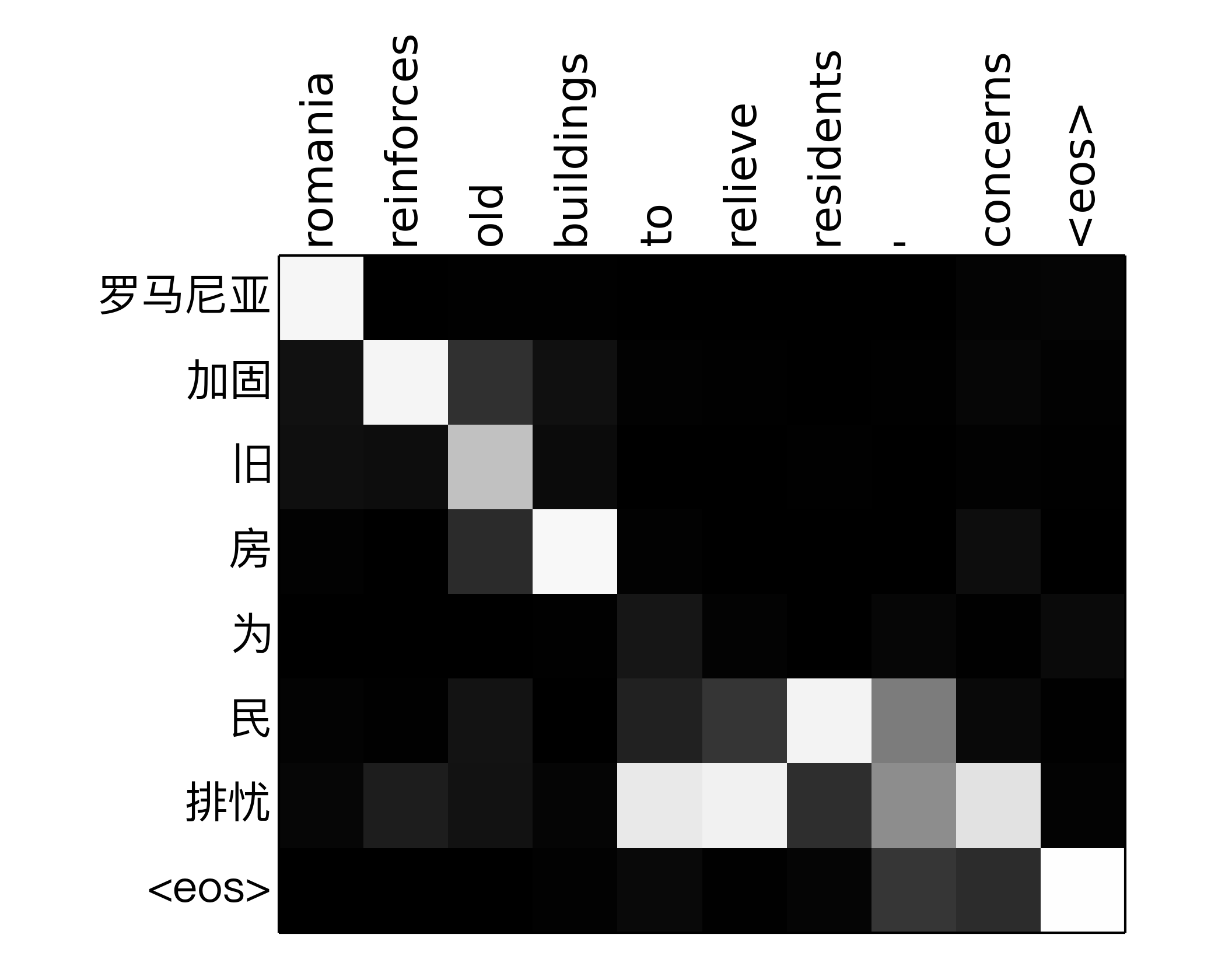}}
\end{center}
\caption{Example alignments. Using coverage mechanism, translated source words are less likely to contribute to generation of the target words next (e.g., top-right corner for the first four Chinese words.).}
\label{figure-alignment-examples}
\end{figure*}

\paragraph{Subjective Evaluation} We also conduct a subjective evaluation to validate the benefit of incorporating coverage. Two human evaluators are asked to evaluate the translations of 200 source sentences randomly sampled from the test sets without knowing from which system a translation is selected. Table~\ref{table-subjective-evaluation} shows the results of subjective evaluation on translation adequacy and fluency.\footnote{Fluency measures whether the translation is fluent, while adequacy measures whether the translation is faithful to the original sentence~\cite{Snover:2009:WMT}.}
GroudHog has a low adequacy since 25.0\% of the source words are under-translated. This is mainly due to the serious under-translation problems on long sentences that consist of several sub-sentences, in which some sub-sentences are completely ignored.
Incorporating coverage significantly alleviates these problems, and reduces 33.2\% and 40.0\% of under-translation and over-translation errors respectively.
Benefiting from this, coverage model improves both translation adequacy and fluency by around 0.2 points.

\subsection{Alignment Quality}
\label{sec-alignment-quality}

\begin{table}[t]
\centering
\begin{tabular}{l|cc}
    {\bf System}					&	SAER &  AER\\
    \hline
    GroundHog					&	67.00	&	54.67\\
    \hdashline
    + Ling. cov. w/o fertility		&	66.75	&	53.55\\
    + Ling. cov. w/ fertility		&	64.85	&	52.13\\
    \hdashline
    + NN cov. w/o gating {\small ($d=1$)} 		&	67.10	&	54.46\\
    + NN cov. w/ gating {\small ($d=1$)}                  &	66.30	&	53.51\\
    + NN cov. w/ gating {\small ($d=10$)}                  &	{\bf 64.25}	&	{\bf 50.50}\\
\end{tabular}
\caption{Evaluation of alignment quality. The lower the score, the better the alignment quality.}
\label{table-alignment-results}
\end{table}

Table~\ref{table-alignment-results} lists the alignment performances.
We find that coverage information improves attention model as expected by maintaining an annotation summarizing attention history on each source word. More specifically, linguistic coverage with fertility significantly reduces alignment errors under both metrics, in which fertility plays an important role. 
NN-based coverages, however, does not significantly reduce alignment errors until increasing the coverage dimension from 1 to 10.
It indicates that NN-based models need slightly more dimensions to encode the coverage information.

Figure~\ref{figure-alignment-examples} shows an example. The coverage mechanism does meet the expectation: the alignments are more concentrated and most importantly, translated source words are less likely to get involved in generation of the target words next.
For example, the first four Chinese words are assigned lower alignment probabilities (i.e., darker color) after the corresponding translation ``\emph{romania reinforces old buildings}'' is produced.

\subsection{Effects on Long Sentences}

\begin{figure*}[t]
\begin{center}
            \includegraphics[width=0.4\textwidth]{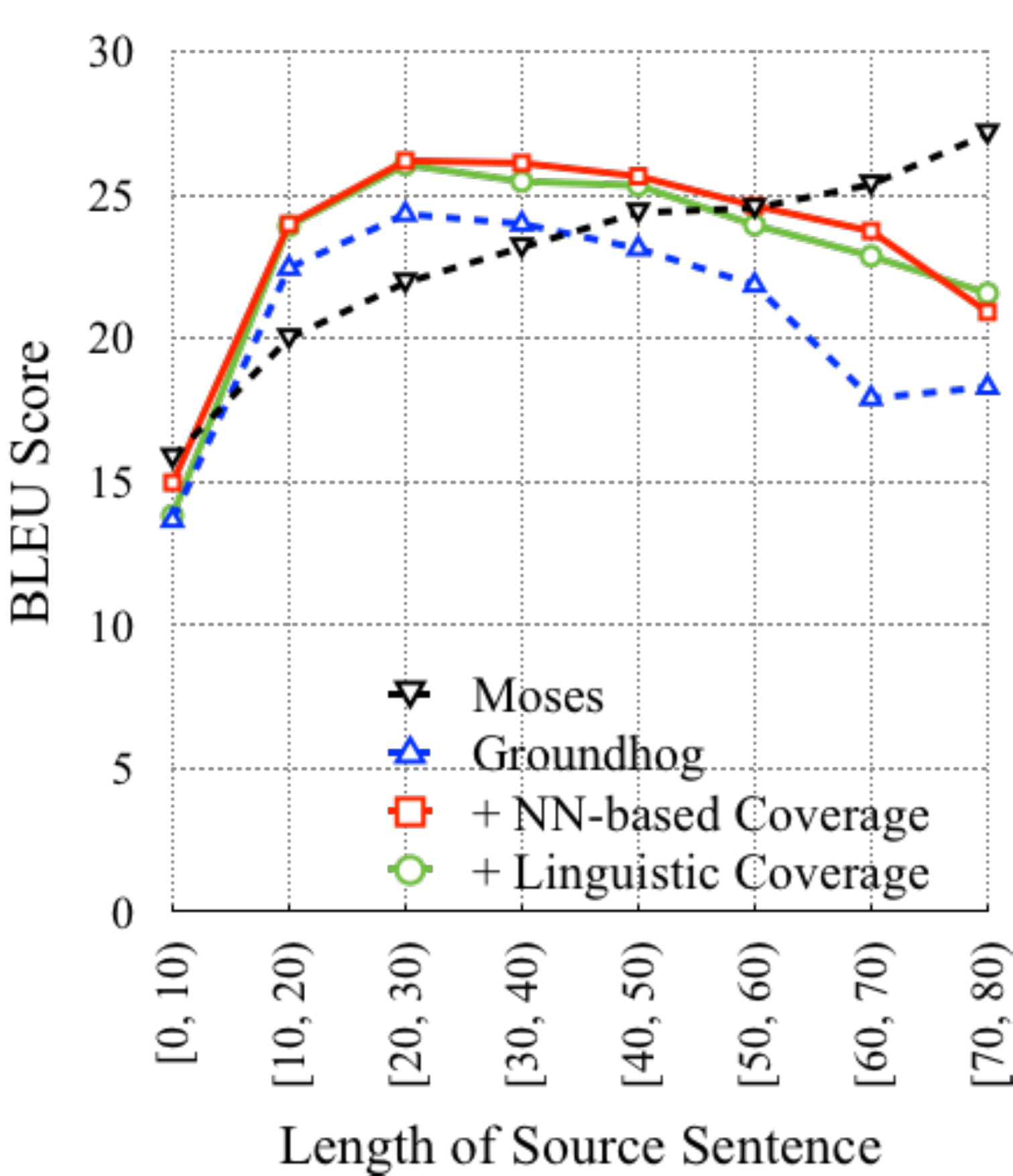}\hspace{0.1\textwidth}
            \includegraphics[width=0.4\textwidth]{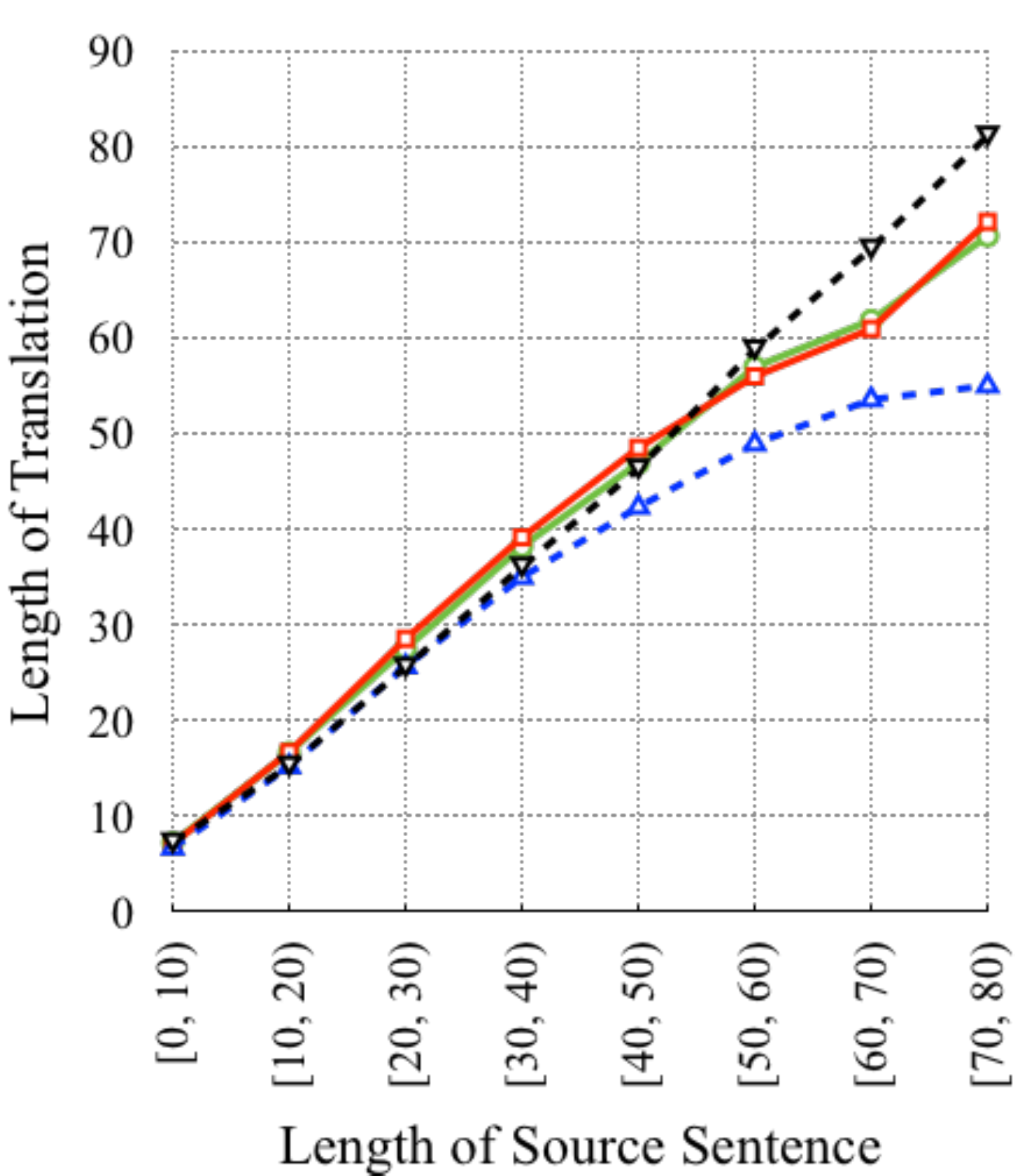}\\
      \caption{Performance of the generated translations with respect to the lengths of the input sentences. Coverage models alleviate under-translation by producing longer translations on long sentences.}
    \label{figure-sentence-len}
  \end{center}
\end{figure*}

Following Bahdanau et al.~\shortcite{Bahdanau:2015:ICLR}, we group sentences of similar lengths together and compute BLEU score and averaged length of translation for each group, as shown in Figure~\ref{figure-sentence-len}. Cho et al.~\shortcite{Cho:2014:SSST} show that the performance of Groundhog drops rapidly when the length of input sentence increases. Our results confirm these findings. One main reason is that Groundhog produces much shorter translations on longer sentences (e.g., $>40$, see right panel in Figure~\ref{figure-sentence-len}), and thus faces a serious under-translation problem. 
\textsc{NMT-Coverage} alleviates this problem by incorporating  coverage information into the attention model, which in general pushes the attention to untranslated parts of the source sentence and implicitly discourages early stop of decoding.
It is worthy to emphasize that both NN-based coverages (with gating, $d=10$) and linguistic coverages (with fertility) achieve similar performances on long sentences, reconfirming our claim that the two variants improve the attention model in their own ways.

As an example, consider this source sentence in the test set:
\begin{quote}
\emph{qi{\'a}od{\=a}n  b{\v e}n  s{\`a}ij{\`\i}  p{\'\i}ngj{\=u}n  d{\' e}f{\=e}n  24.3f{\=e}n , t{\= a}  z{\`a}i  s{\=a}n  zh{\=o}u  qi{\' a}n  ji{\= e}sh{\`o}u  sh{\v o}ush{\`u} \underline{, qi{\' u}du{\`\i}  z{\`a}i  c{\v\i}  q{\=\i}ji{\=a}n  4 sh{\` e}ng  8 f{\` u}}  .}
\end{quote}
Groundhog translates this sentence into:
\begin{quote}
	\emph{jordan achieved an average score of eight weeks ahead with a surgical operation three weeks ago .}
\end{quote}
in which the sub-sentence ``\emph{, qi{\' u}du{\`\i}  z{\`a}i  c{\v\i}  q{\=\i}ji{\=a}n  4 sh{\` e}ng  8 f{\` u}}'' is under-translated.
With the (NN-based) coverage mechanism, \textsc{NMT-Coverage} translates it into:
\begin{quote}
	\emph{jordan 's average score points to UNK this year . he received surgery before three weeks \underline{, with a team in the period} \underline{of 4 to 8} .}
\end{quote}
in which the under-translation is rectified.

The quantitative and qualitative results show that the coverage models indeed help to alleviate under-translation, especially for long sentences consisting of several sub-sentences.

\section{Related Work}

Our work is inspired by recent works on improving attention-based NMT with techniques that have been successfully applied to SMT.
Following the success of Minimum Risk Training (MRT) in SMT~\cite{Och:2003b}, Shen et al.~\shortcite{Shen:2016:ACL} proposed MRT for end-to-end NMT to optimize model parameters directly with respect to evaluation metrics.
Based on the observation that attention-based NMT only captures partial aspects of attentional regularities,
Cheng et al.~\shortcite{Cheng:2016:IJCAI} proposed agreement-based learning~\cite{Liang:2006:NAACL} to encourage bidirectional attention models to agree on parameterized alignment matrices. Along the same direction, inspired by the coverage mechanism in SMT, we propose a coverage-based approach to NMT to alleviate the over-translation and under-translation problems.

Independent from our work, Cohn et al.~\shortcite{Cohn:2016:NAACL} and Feng et al.~\shortcite{Feng:2016:arXiv} made use of the concept of ``fertility'' for the attention model, which is similar in spirit to our method for building the linguistically inspired coverage with fertility.
Cohn et al.~\shortcite{Cohn:2016:NAACL} introduced a feature-based fertility that includes the total alignment scores for the surrounding source words. In contrast, we make prediction of fertility before decoding, which works as a normalizer to better estimate the coverage ratio of each source word.
Feng et al.~\shortcite{Feng:2016:arXiv} used the previous attentional context to represent \emph{implicit fertility} and passed it to the attention model, which is in essence similar to the input-feed method proposed in~\cite{Luong:2015:EMNLP}. Comparatively, we predict \emph{explicit fertility} for each source word based on its encoding annotation, and incorporate it into the linguistic-inspired coverage for attention model. 

\section{Conclusion}

We have presented an approach for enhancing NMT, which maintains and utilizes a coverage vector to indicate whether each source word is translated or not. By encouraging NMT to pay less attention to translated words and more attention to untranslated words, our approach alleviates the serious over-translation and under-translation problems that traditional attention-based NMT suffers from. 
We propose two variants of coverage models: \emph{linguistic coverage} that leverages more linguistic information and \emph{NN-based coverage} that resorts to the flexibility of neural network approximation .
Experimental results show that
both variants achieve significant improvements in terms of translation quality and alignment quality over NMT without coverage.

\section*{Acknowledgement}

This work is supported by China National 973 project 2014CB340301.
Yang Liu is supported by the National Natural Science Foundation of China (No. 61522204) and the 863 Program (2015AA011808).
We thank the anonymous reviewers for their insightful comments.

\balance
\bibliography{all}
\bibliographystyle{acl2016}

\end{document}